\documentclass{article}

\usepackage{microtype}
\usepackage{graphicx}
\usepackage{subfig}
\usepackage{booktabs} 
\usepackage{amsmath}
\usepackage{amsfonts}

\usepackage{xcolor}

\usepackage{hyperref}

\usepackage[accepted]{icml2021}

\icmltitlerunning{Data Augmentation for Meta-Learning}

\begin{document}

\twocolumn[
\icmltitle{Data Augmentation for Meta-Learning}



\icmlsetsymbol{equal}{*}

\begin{icmlauthorlist}
\icmlauthor{Renkun Ni}{umd}
\icmlauthor{Micah Goldblum}{umd}
\icmlauthor{Amr Sharaf}{ms}
\icmlauthor{Kezhi Kong}{umd}
\icmlauthor{Tom Goldstein}{umd}
\end{icmlauthorlist}

\icmlaffiliation{umd}{Department of Computer Science, University of Maryland, College Park}
\icmlaffiliation{ms}{Microsoft}

\icmlcorrespondingauthor{Renkun Ni}{rn9zm@cs.umd.edu}

\icmlkeywords{Machine Learning, ICML}

\vskip 0.3in
]



\printAffiliationsAndNotice{}  

\begin{abstract}
Conventional image classifiers are trained by randomly sampling mini-batches of images. To achieve state-of-the-art performance, practitioners use sophisticated data augmentation schemes to expand the amount of training data available for sampling. In contrast, meta-learning algorithms sample support data, query data, and tasks on each training step. In this complex sampling scenario, data augmentation can be used not only to expand the number of images available per class, but also to generate entirely new classes/tasks. We systematically dissect the meta-learning pipeline and investigate the distinct ways in which data augmentation can be integrated at both the image and class levels. Our proposed meta-specific data augmentation significantly improves the performance of meta-learners on few-shot classification benchmarks.
\end{abstract}

\section{Introduction}
\label{Introduction}

\par Data augmentation has become an essential part of the training pipeline for image classifiers and similar systems, as it offers a simple and efficient way to significantly improve performance \citep{cubuk2018autoaugment, zhang2017mixup}.  In contrast, little work exists on data augmentation for meta-learning. Existing frameworks for few-shot image classification use only horizontal flips, random crops, and color jitter to augment images in a way that parallels augmentation for conventional training \citep{bertinetto2018meta, lee2019meta}.  Meanwhile, meta-learning methods have received increasing attention as they have reached the cutting edge of few-shot performance.  While new meta-learning algorithms emerge at a rapid rate, we show that, like image classifiers, meta-learners can achieve significant performance boosts through carefully chosen data augmentation strategies that are injected into various stages of the meta-learning pipeline.

Meta-learning frameworks use data for multiple purposes during each gradient update, which creates the possibility for a diverse range of data augmentations that are not possible within the standard training pipeline. At the same time, it is still unclear how different categories of data within the training pipeline impact meta-learning performance. We explore these possibilities and discover combinations of augmentation types that improve performance over existing methods. Our contributions can be summarized as follows:
\begin{itemize}
\item First, we break down the meta-learning pipeline and find that each component contributes differently to meta-learning performance: meta-learners are very sensitive to the amount of query data and number of tasks and less sensitive to the amount of support data.
\item Based on these findings, we uncover four modes of augmentations for meta-learning that differ in where in the training pipeline they are applied: support augmentation, query augmentation, task augmentation, and shot augmentation.
\item We test these four modes using a pool of image augmentations, and we confirm that query augmentation is critical, while support augmentation often does not provide performance benefits and may even degrade accuracy in some cases.
\item Finally, we combine augmentations and implement a MaxUp strategy, which we call Meta-MaxUp, to maximize performance.  We achieve significant performance boosts for popular meta-learners on few-shot benchmarks such as mini-ImageNet, CIFAR-FS and Meta-Dataset.
\end{itemize}

\section{Background and Related Work}
\subsection{The Meta-Learning Framework}
Meta-learning algorithms aim to learn a network that can easily adapt to new tasks with limited data and generalize to unseen examples. In order to achieve this, they simulate the adaptation and evaluation procedure during meta-training.  To simulate an $N$-way classification task, $\mathcal{T}_i$, we sample \emph{support} data $\mathcal{T}^s_i$ and \emph{query} data $\mathcal{T}^q_i$, so that $\mathcal{T}_i = \{\mathcal{T}^s_i, \mathcal{T}^q_i\}$.  As we will detail in the following paragraph, support will be used to simulate few-shot training data, while query will be used to simulate unseen testing data.  Note that \emph{shot} denotes the number of training samples per class available for fine-tuning on a given task during the testing phase.
%
%

Adopting common terminology from the literature, the archetypal meta-learning algorithm contains an \emph{inner loop} and an \emph{outer loop} in each parameter update of the training procedure. In the inner loop, a model is first fine-tuned or adapted on support data $\mathcal{T}^s_i$. 
Then, in the outer loop, the updated model is evaluated on query data $\mathcal{T}^q_i$, and minimizes loss on the query data with respect to the model's parameters before fine-tuning. This loss minimization step may require computing the gradient through the fine-tuning procedure.  Existing meta-learning algorithms apply various methods for fine-tuning on support data during the inner loop. Some algorithms, such as MAML and Reptile~\citep{finn2017model, nichol2018first}, update all the parameters in the network using gradient descent during fine-tuning on support data. Other algorithms, such as MetaOptNet and R2-D2~\citep{lee2019meta, bertinetto2018meta}, only update the parameters from the linear classifier layer during the fine-tuning while keeping the feature extraction layers frozen. These methods benefit from the simplicity and the convexity of the inner loop optimization problem. Similarly, metric learning approaches, such as~\citep{snell2017prototypical, kye2020transductive}, freeze the feature extraction layers as well, and create class centroids from the support data during the inner loop. These method have low cost training iterations, and can be applied on deeper architectures to achieve better performance. In this work, we mainly focus on the latter algorithms due to their stronger performance. Further details of the algorithms used in our experiments can be found in Section~\ref{sec:setup}. 

\subsection{Preventing Overfitting in Meta-Learning}
Meta-learners are known to be particularly vulnerable to overfitting~\citep{rajendran2020meta}. One work, MetaMix, proposes averaging support and query features to prevent the model from memorizing the query data and ignoring support~\citep{yao2020don}. Recently, another work adds random noise to the label space to make the model rely on support data~\citep{rajendran2020meta}. In the context of few-shot classification, random shuffling labels within tasks alleviates this kind of overfitting and is commonplace in meta-learning algorithms~\citep{yin2019meta, rajendran2020meta}. However, as shown in Figure~\ref{fig:trainval}, overfitting to training tasks remains a problem. One recent work has developed a data augmentation method to overcome this problem \citep{liu2020task}.  This method simply rotates all images in a class by a large degree and considers this new rotated class distinct from its parent class. This effectively increases the number of possible few-shot tasks that can be sampled during training.

A different line of work instead applies regularizers to prevent overfitting and improve few-shot classification \citep{yin2019meta, goldblum2020unraveling}. Yet additional work has developed methods for labeling and augmenting unlabeled data \citep{antoniou2019assume, chen2019image}, generative models for deforming images in one-shot metric learning \citep{chen2019image2}, and feature space data augmentation for adapting language models to new unseen intents \citep{kumar2019closer}.

\subsection{Few-shot Benchmarks}
\label{sec:benchmark}
In this paper, we perform our experiments on the mini-ImageNet and CIFAR-FS datasets as well as the Meta-Dataset benchmark~\citep{vinyals2016matching, bertinetto2018meta, triantafillou2019meta}.  Mini-ImageNet is a few-shot learning dataset derived from the ImageNet classification dataset~\citep{deng2009imagenet}, and CIFAR-FS is derived from CIFAR-100~\citep{krizhevsky2009learning}. Each of these datasets contains 64 training classes, 16 validation classes, and 20 classes for testing. In each class, there are 600 images, and both Mini-ImageNet and CIFAR-FS have 60000 images in total. Meta-Dataset is a large-scale diverse benchmark consisting of 10 different image classification subdatasets with distinct data distributions. This diversity allows us to measure cross-domain generalization. 

\section{The Anatomy of Data Augmentation for Meta-Learning}
\label{Anatomy}
\subsection{Where Does Dataset Diversity Matter Most? In the Support, Query or Tasks?}
\label{data_size}

Since data augmentation techniques aim to increase the amount of training samples, learning algorithms that are sensitive to the amount of training data may benefit more from these techniques.  In this section, before we introduce data augmentations, we investigate how sensitive meta-learning algorithms are to the amount of support data, query data, and tasks. Typically, support and query data are sampled from the same pool (the entire training set).  

To examine the impact of dataset diversity on various stages of meta-learning, we perform an ablation where we limit the diversity of each stage.
We first reduce the pool of support data to a fixed subset of only five independent samples per class while sampling query data from the entire training set.  That is, whenever a support image is sample from class $c$, it is only sampled from the five-image subset associated with that class instead of from all training data in that class. Interestingly, we find that test accuracy remains almost the same as baseline performance (see Table \ref{tab:data_size}).  In fact, if we replace those five support images per class with fixed random noise images, we still only observe a small degradation in performance.  We then instead shrink the pool of query data (but not support), and we see a much larger decrease in test accuracy.  These experiments suggest that meta-learning is fairly insensitive to the amount and quality of support but not query data.  This observation agrees with our following finding that augmenting query data is far more beneficial than augmenting support.

Since we also consider task-level augmentation, we now examine how sensitive meta-learning is to a decrease in task diversity. As CIFAR-FS contains $64$ training classes, there are ${64 \choose 5} = 7624512$  5-way classification problems that can be sampled during each iteration of meta-learning.  We reduce the number of tasks by randomly batching classes into just $13$ distinct 5-way classification tasks before training, and we only train on these 13 tasks.  We do this in such a way that all classes, and therefore training data, are used during training.  We observe that this process noticeably degrades test accuracy, and we conclude that there may be room to improve performance by augmenting the number of tasks (see Table \ref{tab:data_size}). To verify that this impact of dataset diversity generalizes, we run additional experiments on Mini-ImageNet and with other backbones. The results are shown in Appendix A, and these experiments support the aforementioned findings as well.

\begin{table}[h!]
\caption{Few-shot classification accuracy (\%) using R2-D2 and a ResNet-12 backbone for various data size manipulations on CIFAR-FS. ``Support'', ``Query'' and ``Task'' columns denote the number of samples per class for support and query data and the number of total tasks available for sampling.  The first row contains baseline performance.  Confidence intervals have radius equal to one standard error.}
\begin{center}
\scalebox{0.9}{
\begin{tabular}{cccccc}
\hline
Support & Query & Task & 1-shot & 5-shot \\ \hline
600 & 600 & full & 71.73 $\pm$ 0.37 & 84.39 $\pm$ 0.25 \\ \hline
5 & 600 & full & 70.97 $\pm$ 0.36 & 84.51 $\pm$ 0.24 \\ \hline
5 (random) & 600 & full & 58.15 $\pm$ 0.36 & 76.26 $\pm$ 0.27\\ \hline
600 & 5 & full & 60.25 $\pm$ 0.37 & 77.05 $\pm$ 0.28 \\ \hline
600 & 600 & 13 & 68.24 $\pm$ 0.38 & 81.77 $\pm$ 0.26 \\ \hline
\end{tabular}
}
\end{center}
\label{tab:data_size}
\vskip -0.1in
\end{table}

\subsection{Data Augmentation Modes}
\label{modes}
Motivated by the observation that meta-learning is more sensitive to the amount of query data and tasks than support, we delineate four modes of data augmentation for meta-learning which may be employed individually or combined.

\paragraph{Support augmentation:} Data augmentation may be applied to support data in the inner loop of fine-tuning.  This strategy enlarges the pool of fine-tuning data.

\paragraph{Query augmentation:} Data augmentation alternatively may be applied to query data.  This strategy enlarges the pool of evaluation data to be sampled during training.

\paragraph{Task augmentation:} We can increase the number of possible tasks by uniformly augmenting whole classes to add new classes with which to train.  For example, a vertical flip applied to all car images yields a new upside-down car class which may be sampled during training.

\paragraph{Shot augmentation:}  At test time, we can artificially amplify the shot by adding additional augmented copies of each image. Shot augmentation can also be used during training by adding copies of each support image via augmentation.  Shot augmentation during training may be needed to prepare a network for the use of test-time shot augmentation.

Existing meta-learning algorithms for few-shot image classification typically apply standard augmentations (horizontal flips, random crops, and color jitter) on all images that come from the data loader without considering the purpose of each image.  As a result, the same augmentation occurs on both support and query images~\citep{gidaris2018dynamic, qiao2018few}. In Section \ref{Experiments}, we test the four modes of data augmentation enumerated above in isolation across a large array of specific augmentations. We find that query augmentation is far more critical than support augmentation for increasing performance.  In fact, support augmentation often hurts performance.  Additionally, we find that task augmentation, when combined with query augmentation, can offer further boosts in performance when compared with existing frameworks.

\subsection{Data Augmentation Techniques}

For each of the data augmentation modes described above, we try a variety of specific data augmentation techniques. Some techniques are only applicable to support, query, and shot modes or solely to the task mode. We use an array of standard augmentation techniques as well as CutMix  \citep{yun2019cutmix}, MixUp \citep{zhang2017mixup}, and Self-Mix \citep{seo2020self}. In the context of the task augmentation mode, we apply these the same way to every image in a class in order to augment the number of classes. For example, we use MixUp to create a half-dog-half-truck class where every image is the average of a dog image and a truck image. We also try combining multiple classes into one class as a task augmentation mode. 

In general, techniques that greatly change the image distribution (i.e. a vertical flip, which does not naturally appear in the dataset) are better suited for task augmentations while techniques that preserve the image distribution (e.g., random crops, which produce images that are presumably within the support of the image distribution) are typically better suited for the support, query, and shot augmentation modes. The baseline models we compare to use horizontal flip, random crop, and color jitter augmentation techniques at both the support and query levels since this combination is prevalent in the literature. More details on our pool of augmentation techniques can be found in Appendix B.

\subsection{Meta-MaxUp Augmentation for Meta-Learning}

Recent work proposes MaxUp augmentation to alleviate overfitting during the training of classifiers \citep{gong2020maxup}. This strategy applies many augmentations to each image and chooses the augmented image which yields the highest loss. MaxUp is conceptually similar to adversarial training \citep{madry2019deep}. Like adversarial training, MaxUp involves solving a saddlepoint problem in which loss is minimized with respect to parameters while being maximized with respect to the input. In the standard image classification setting, MaxUp, together with CutMix, improves generalization and achieves state-of-the-art performance on ImageNet. Here, we extend MaxUp to the setting of meta-learning. Before training, we select a pool, $\mathcal{S}$, of data augmentations from the four modes as well as their combinations. For example, $\mathcal{S}$ may contain horizontal flip shot augmentation, query CutMix, and the combination of both.  During each iteration of training, we first sample a batch of tasks, each containing support and query data, as is typical in the meta-learning framework. For each element in the batch, we randomly select $m$ augmentations from the set $\mathcal{S}$, and we apply these to the task, generating $m$ augmented tasks with augmented support and query data. Then, for each element of the batch of tasks originally sampled, we choose the augmented task that maximizes loss, and we perform a parameter update step to minimize training loss. Formally, we solve the minimax optimization problem,
\begin{equation*}
    \min_{\theta} \mathbb{E}_{\mathcal{T}} \Big[\max_{M\in\mathcal{S}}\mathcal{L}(F_{\theta'}, M(\mathcal{T}^q))\Big],
\end{equation*}
where $\theta' = \mathcal{A}(\theta, M(\mathcal{T}^s))$, $\mathcal{A}$ denotes fine-tuning, $F$ is the base model with parameters $\theta$, $\mathcal{L}$ is the loss function used in the outer loop of training, and $\mathcal{T}$ is a task with support and query data $\mathcal{T}^s$ and $\mathcal{T}^q$, respectively.  Algorithm \ref{alg:Meta-MaxUp} contains a more thorough description of this pipeline in practice (adapted from the standard meta-learning algorithm in \citet{goldblum2019adversarially}).

\begin{algorithm}[h]
\caption{Meta-MaxUp}
\label{alg:Meta-MaxUp}
\begin{algorithmic}
\STATE {\bfseries Require:} Base model $F_\theta$, fine-tuning algorithm $\mathcal{A}$, learning rate $\gamma$, set of augmentations $\mathcal{S}$, and distribution over tasks $p(\mathcal{T})$.\\
\STATE Initialize $\theta$, the weights of $F$; \\
\WHILE{not done}
\STATE Sample batch of tasks, $\{\mathcal{T}_i\}_{i=1}^n$, where $\mathcal{T}_i \sim p(\mathcal{T})$ and $\mathcal{T}_i = (\mathcal{T}_i^s, \mathcal{T}_i^q)$. \\
\FOR{$i=1,...,n$}
\STATE Sample $m$ augmentations, $\{M_j\}_{j=1}^m$, from $\mathcal{S}$.\\
\STATE Compute $k=\operatorname*{argmax}_{j} \mathcal{L}(F_{\theta_j}, M_j(\mathcal{T}_i^{q}))$, where $\theta_{j} = \mathcal{A}(\theta, M_j(\mathcal{T}_i^s))$.
\STATE Compute gradient $g_i = \nabla_{\theta} \mathcal{L}(F_{\theta_{k}}, M_k(\mathcal{T}_i^{q}))$.
\ENDFOR
\STATE Update base model parameters: $\theta \leftarrow \theta - \frac{\gamma}{n} \sum_i g_i$.
\ENDWHILE
\end{algorithmic}
\end{algorithm}

\section{Experiments}
\label{Experiments}

In this section, we empirically demonstrate the following:
\begin{enumerate}
    \item Augmentations applied in the four distinct modes behave differently. In particular, query and task augmentation are far more important than support augmentation. (Section \ref{sec:experiments:individual})
    \item Meta-specific data augmentation strategies can improve performance over the generic strategies commonly used for meta-learning. (Section \ref{sec:experiments:combine})
    \item We further boost performance by combining augmentations with Meta-MaxUp. (Section \ref{sec:experiment:metamaxup})
    \item Our proposed augmentation Meta-MaxUp greatly improves performance on cross-domain benchmarks as well. (Section \ref{sec:experiment:metadataset})
\end{enumerate}

\subsection{Experimental Setup}
\label{sec:setup}

We conduct experiments on four meta-learning algorithms: ProtoNet~\citep{snell2017prototypical}, R2-D2~\citep{bertinetto2018meta}, MetaOptNet~\citep{lee2019meta}, and MCT~\citep{kye2020transductive}.  ProtoNet is a metric-learning method that uses a prototype learning head, which classifies samples by extracting a feature vector and then performing a nearest-neighbor search for the closest class prototype.  R2-D2 and MetaOptNet instead use differentiable solvers with a ridge regression and SVM head, respectively.  These methods extract feature vectors and then apply a standard linear classifer to assign class labels.  MCT improves upon ProtoNet by meta-learning confidence scores.  We experiment with all of these different classifier head options, all using the ResNet-12 backbone proposed by \citet{oreshkin2018tadam} as well as the four-layer convolutional architectures proposed by \citet{snell2017prototypical} and \citet{bertinetto2018meta}.

We perform our experiments on the aforementioned benchmark datasets, mini-ImageNet, CIFAR-FS, and Meta-Dataset. A description of training hyperparameters and computational complexity can be found in Appendix C. 
We report confidence intervals with a radius of one standard error.

Few-shot learning may be performed in either the inductive or transductive setting.  Inductive learning is a standard method in which each test image is evaluated separately and independently.  In contrast, transduction is a mode of inference in which the few-shot learner has access to all unlabeled testing data at once and therefore has the ability to perform semi-supervised learning by training on the unlabelled data.  For fair comparison, we only compare inductive methods to other inductive methods. A PyTorch implementation of our data augmentation methods for meta-learning can be found at: \texttt{
https://github.com/RenkunNi/MetaAug}

\subsection{An Empirical Comparison of Augmentation Modes}\label{sec:aug_mode}
\label{sec:experiments:individual}

We empirically evaluate the performance of all four different augmentation modes identified in Section \ref{modes} on the CIFAR-FS dataset using an R2-D2 base-learner paired with both a 4-layer convolutional network backbone (as used in the original work~\citep{bertinetto2018meta}) and a ResNet-12 backbone. We report the results of the most effective augmentations for each mode on the ResNet-12 backbone in Table~\ref{tab:active_pool}. 
Appendix D contains an extensive table with various augmentations and both backbones.

Table~\ref{tab:active_pool} demonstrates that each mode of augmentation individually can improve performance.  Augmentation applied to query data is consistently more effective than the other augmentation modes.  In particular, simply applying CutMix to query samples improves accuracy by as much as 3\% on both backbones.  In contrast, most augmentations on support data actually damage performance.  The overarching conclusion of these experiments is that the four modes of data augmentation for meta-learning behave differently.  Existing meta-learning methods, which apply the same augmentations to query and support data without using task and shot augmentation, may be achieving suboptimal performance. 

\begin{table}[!h]
  \caption{Few-shot classification accuracy (\%) using R2-D2 and a ResNet-12 backbone on the CIFAR-FS dataset with the most effective data augmentations for each mode shown. Confidence intervals have radii equal to one standard error. Best performance in each category is bolded. Query CutMix is consistently the most effective single augmentation for meta-learning.}
  \vskip 0.15in
  \centering
  \begin{tabular}{@{}lccc@{}}
    \toprule
    Method & Mode & 1-shot & 5-shot \\
    \midrule
    Baseline & - &  71.95 $\pm$ 0.37 & 84.56 $\pm$ 0.25  \\
    \midrule
    CutMix & Support &  72.79 $\pm$ 0.37  & 84.70 $\pm$ 0.25 \\
    Self-Mix & Support & 71.96 $\pm$ 0.36 &  84.84 $\pm$ 0.25 \\
    \midrule
    CutMix & Query & \textbf{75.97 $\pm$ 0.34} & \textbf{87.28 $\pm$ 0.23} \\
    Self-Mix & Query & 73.59 $\pm$ 0.35 & 86.14 $\pm$ 0.24 \\
    \midrule
    Large Rotation & Task & 73.79 $\pm$ 0.36 & 85.81 $\pm$ 0.24 \\
    MixUp & Task & 72.05 $\pm$ 0.37 & 85.27 $\pm$ 0.25 \\
    \midrule
    Random Crop & Shot & 70.56 $\pm$ 0.37  & 83.87 $\pm$ 0.25 \\
    Horizontal Flip & Shot & 73.25 $\pm$ 0.36 & 85.06 $\pm$ 0.25 \\
    \bottomrule
  \end{tabular}
  \label{tab:active_pool}
  \vskip -0.1in
\end{table}

\subsection{Combining Augmentations}
\label{sec:experiments:combine}

After studying each mode of data augmentation individually, we combine augmentations in order to find out how augmentations interact with each other.  We build on top of query CutMix since this augmentation was the most effective in the previous section.  We combine query CutMix with other effective augmentations from Table~\ref{tab:active_pool}, and we conduct experiments on the same backbones and dataset.  Results on the ResNet-12 backbone are reported in Table~\ref{tab:combination}, and a full table with additional results can be find in Appendix E. 
Interestingly, when we use CutMix on both support and query images, we observe worse performance than simply using CutMix on query data alone.  Again, this demonstrates that meta-learning demands a careful and meta-specific data augmentation strategy.  In order to further boost performance, we will need an intelligent method for combining various augmentations.  We propose Meta-MaxUp as this method.

\begin{table}[t]
  \centering
  \caption{Few-shot classification accuracy (\%) using R2-D2 and a ResNet-12 backbone on the CIFAR-FS dataset with combinations of augmentations and query CutMix. ``S",``Q",``T" denote ``Support", ``Query", and ``Task" modes, respectively.  While adding augmentations can help, it can also hurt, so additional augmentations must be chosen carefully.}
  \vskip 0.15in
  \begin{tabular}{@{}lcc@{}}
    \toprule
    Mode & 1-shot & 5-shot \\
    \midrule
    CutMix & 75.97 $\pm$ 0.34  & 87.28 $\pm$ 0.23  \\
    + CutMix (S) & 75.00 $\pm$ 0.37 & 85.37 $\pm$ 0.25 \\
    + Random Erase (S) & 75.84 $\pm$ 0.34 & 87.19 $\pm$ 0.24 \\
    + Random Erase (Q) & 75.08 $\pm$ 0.35 & 87.14 $\pm$ 0.23  \\
   + Self-Mix (S) & \textbf{76.27 $\pm$ 0.34} & 87.52 $\pm$ 0.24 \\
    + Self-Mix (Q) & 76.04 $\pm$ 0.34 & 87.45 $\pm$ 0.24 \\
    + MixUp (T) & 75.97 $\pm$ 0.34 & 86.66 $\pm$ 0.24 \\
    + Rotation (T) & 75.74 $\pm$ 0.34 & \textbf{87.68 $\pm$ 0.24} \\
    + Horizontal Flip (Shot) & 76.23 $\pm$ 0.34  & 87.36 $\pm$ 0.24 \\
    \bottomrule
  \end{tabular}
  \vskip -0.1in
  \label{tab:combination}
\end{table}


\begin{figure*}[!h]
\centering
\includegraphics[scale=0.5]{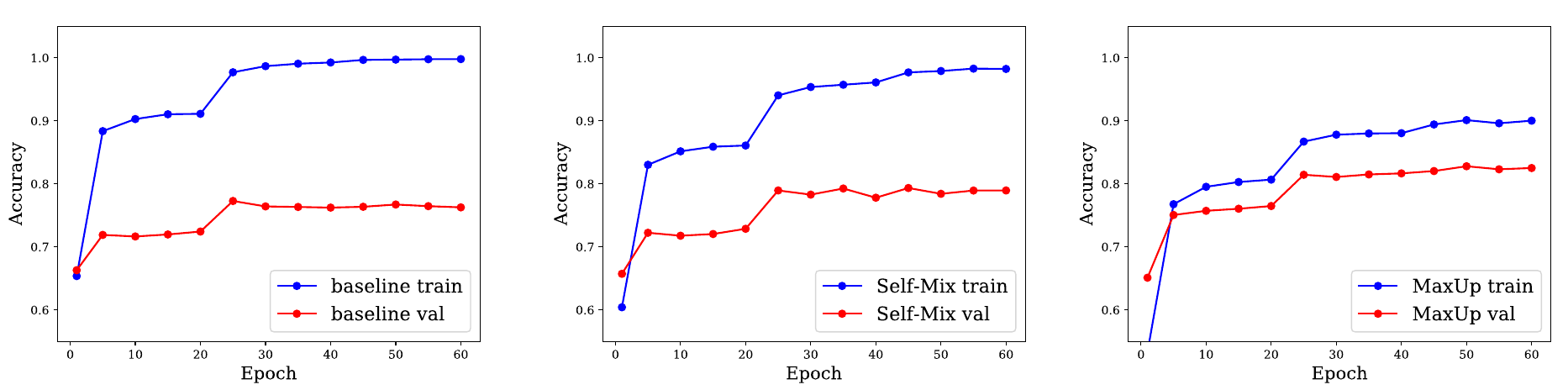}
\caption{Training and validation accuracy for R2-D2 meta-learner with ResNet-12 backbone on the CIFAR-FS dataset. (Left) Baseline model (Middle) query Self-Mix (Right) Meta-MaxUp. Better data augmentation strategies, such as MaxUp, narrow the generalization gap and prevent overfitting.} \label{fig:trainval}
\vskip -0.1in
\end{figure*}

\subsection{Meta-MaxUp Further Improves Performance}
\label{sec:experiment:metamaxup}

In this section, we evaluate our proposed Meta-MaxUp strategy in the same experimental setting as above for various values of $m$ and different data augmentation pool sizes. Table~\ref{tab:maxup} contains the results, and a detailed description of the augmentation pools as well as the full results can be found in Appendix F. 
Rows beginning with ``CutMix'' denote experiments in which the pool of augmentations simply includes many CutMix samples.  ``Single'' denotes experiments in which each augmentation in $\mathcal{S}$ is of a single type, while ``Medium'' and ``Large'' denote experiments in which each element of $\mathcal{S}$ is a combination of augmentations, for example CutMix+rotation.  Combinations greatly expand the number of augmentations in the pool.  Rows with $m=1$ denote experiments where we do not maximize loss in the inner loop and thus simply apply randomly sampled data augmentation for each task.  As we increase $m$ and include a large number of augmentations in the pool, we observe performance boosts as high as $4\%$ over the baseline, which uses horizontal flip, random crop, and color jitter data augmentations from the original work corresponding to the R2-D2 meta-learner~\citep{bertinetto2018meta}.

\begin{table}[!h]
  \caption{Few-shot classification accuracy (\%) using R2-D2 and a ResNet-12 backbone on the CIFAR-FS dataset for Meta-MaxUp over different sizes of augmentation pools and numbers of samples.  As $m$ and the pool size increase, so does performance.  Meta-MaxUp is able to pick effective augmentations from a large pool.}
  \vskip 0.15in
  \centering
  \begin{tabular}{@{}lccc@{}}
    \toprule
    Pool & m & 1-shot & 5-shot \\
    \midrule
    Baseline & - & 71.95 $\pm$ 0.37 & 84.56 $\pm$ 0.25  \\
    \midrule
    CutMix & 1 & 75.97 $\pm$ 0.34 & 87.28 $\pm$ 0.23 \\
    Single & 1 & 75.71 $\pm$ 0.35 & 87.44 $\pm$ 0.43  \\
    Medium & 1 & 75.60 $\pm$ 0.34 & 87.35 $\pm$ 0.23 \\
    Large & 1 & 75.44 $\pm$ 0.34 & 87.47 $\pm$ 0.23  \\
    \midrule
    CutMix & 2 & 74.93 $\pm$ 0.36 & 87.14 $\pm$ 0.24 \\
    Single & 2 & 75.81 $\pm$ 0.34 & 87.33 $\pm$ 0.23 \\
    Medium & 2 & 76.49 $\pm$ 0.33 & 88.20 $\pm$  0.22 \\
    Large & 2 & 76.59 $\pm$ 0.34 & 88.11 $\pm$ 0.23\\
    \midrule
    CutMix & 4 & 75.08 $\pm$ 0.23 &87.60 $\pm$ 0.24 \\
    Single & 4 & 76.82 $\pm$ 0.24 & 88.14 $\pm$ 0.23\\
    Medium & 4 & 76.30 $\pm$ 0.24 & 88.29 $\pm$  0.22\\
    Large & 4 & \textbf{76.99 $\pm$ 0.24} & \textbf{88.35 $\pm$  0.22} \\
    \bottomrule
  \end{tabular}
  \label{tab:maxup}
\end{table}

We explore the training benefits of these meta-specific training schemes by examining saturation during training.  To this end, we plot the training and validation accuracy over time for R2-D2 meta-learners with ResNet-12 backbones using baseline augmentations, query Self-Mix, and Meta-MaxUp with a medium sized pool and $m=4$.  See Figure \ref{fig:trainval} for training and validation accuracy curves.  With only baseline augmentations, validation accuracy stops increasing immediately after the first learning rate decay.  This suggests that baseline augmentations do not prevent overfitting during meta-training. In contrast, we observe that models trained with Meta-MaxUp do not quickly overfit and continue improving validation performance for a greater number of epochs.  Meta-MaxUp visibly reduces the generalization gap.

\begin{figure}[h]
\centering
\includegraphics[scale=0.53]{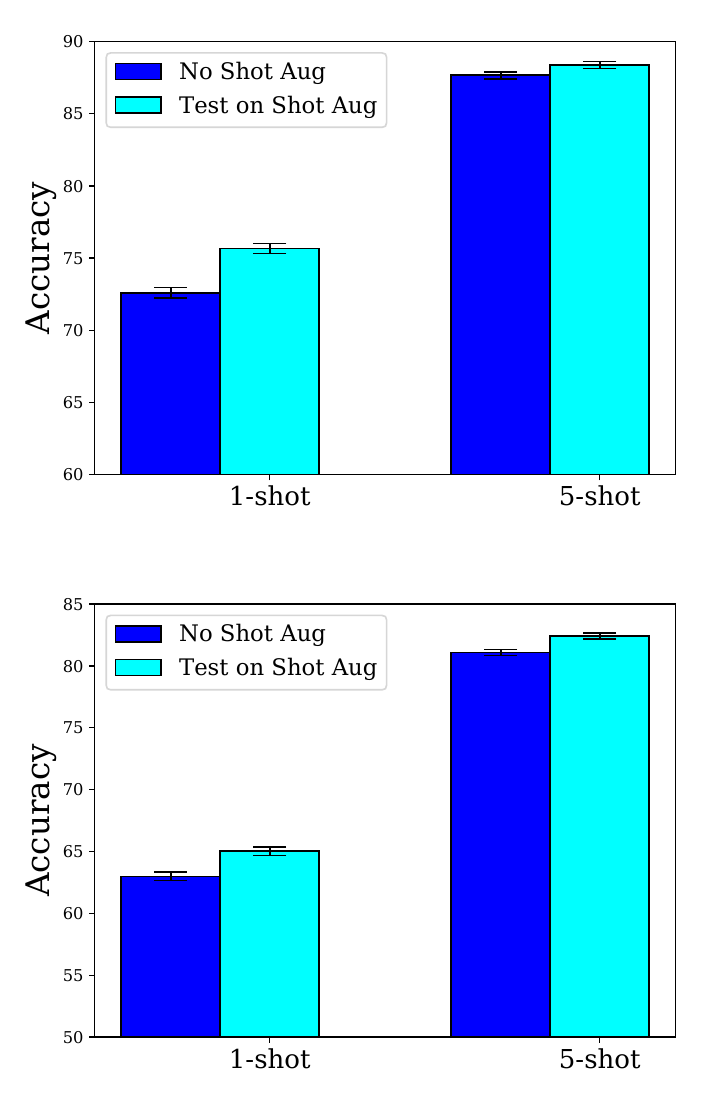}
\caption{Performance with shot augmentation using MetaOptNet trained with the proposed Meta-MaxUp. (Top) 1-shot and 5-shot on CIFAR-FS (Bottom) 1-shot and 5-shot on mini-ImageNet. }\label{fig:shotaug}
\end{figure}

\begin{table*}[h!]
  \caption{Few-shot classification accuracy (\%) on CIFAR-FS and mini-ImageNet. ``+ DA" denotes training with CutMix (Q) + Rotation (T), and ``+ MM" denotes training with Meta-MaxUp. ``CNN-4'' denotes a 4-layer convolutional network with 96, 192, 384, and 512 filters in each layer~\citep{bertinetto2018meta}. ``64-64-64-64'' denotes the 4-layer CNN backbone from \citet{snell2017prototypical}.}
  \vskip 0.15in
  \centering
  \scalebox{0.85}{
  \begin{tabular}{@{}lccccc@{}}
    \toprule
    & &\multicolumn{2}{c}{CIFAR-FS }&\multicolumn{2}{c}{mini-ImageNet}\\
    \cmidrule(r){3-4}\cmidrule(r){5-6}
    Method & Backbone & 1-shot & 5-shot & 1-shot & 5-shot\\
    \midrule
    R2-D2 & CNN-4 & 67.56 $\pm$ 0.35 & 82.39 $\pm$ 0.26 & 56.15 $\pm$ 0.31 & 72.46 $\pm$ 0.26  \\
    $\quad$ + DA & CNN-4 & 70.54 $\pm$  0.33 & 84.69 $\pm$ 0.24 &57.60 $\pm$ 0.32 & 74.69 $\pm$ 0.25\\
    $\quad$+ MM & CNN-4 & \textbf{71.10 $\pm$ 0.34} & \textbf{85.50 $\pm$ 0.24} & \textbf{58.18 $\pm$ 0.32} & \textbf{75.35 $\pm$ 0.25}  \\
    \midrule
    R2-D2 & ResNet-12 & 71.95 $\pm$  0.37 & 84.56 $\pm$  0.25  &60.46 $\pm$  0.32 & 76.88 $\pm$  0.24 \\
    $\quad$+ DA & ResNet-12 & 76.17 $\pm$ 0.34 & 87.74 $\pm$ 0.24 & \textbf{65.54 $\pm$ 0.32} & 81.52 $\pm$ 0.23  \\
    $\quad$+ MM & ResNet-12 &  \textbf{76.65 $\pm$ 0.33} & \textbf{88.57 $\pm$ 0.24} & 65.15 $\pm$ 0.32 & \textbf{81.76 $\pm$ 0.24} \\
    \midrule
    ProtoNet & 64-64-64-64 & 60.91 $\pm$ 0.35 & 79.73 $\pm$ 0.27 & 47.97 $\pm$ 0.32 & 70.13 $\pm$ 0.27 \\
    $\quad$+ DA & 64-64-64-64 & 62.21 $\pm$ 0.36 & 80.70 $\pm$ 0.27 & \textbf{50.38 $\pm$ 0.32} & \textbf{71.44 $\pm$ 0.26} \\
    $\quad$+ MM & 64-64-64-64 & \textbf{63.01 $\pm$ 0.36} & \textbf{80.85 $\pm$ 0.25} & 50.06 $\pm$ 0.32 &  71.13 $\pm$ 0.26\\
    \midrule
     ProtoNet & ResNet-12 & 70.21 $\pm$ 0.36 & 84.26 $\pm$ 0.25 & 57.34 $\pm$ 0.34 &75.81 $\pm$ 0.25 \\
    $\quad$+ DA & ResNet-12 & 74.30 $\pm$ 0.36 & 86.24 $\pm$ 0.24 &60.82 $\pm$ 0.34 & 78.23 $\pm$ 0.25\\
     $\quad$+ MM & ResNet-12 & \textbf{76.05 $\pm$ 0.34} & \textbf{87.84 $\pm$ 0.23} & \textbf{62.81 $\pm$ 0.34} & \textbf{79.38 $\pm$ 0.24}\\
     \midrule
    MetaOptNet & ResNet-12 & 70.99 $\pm$ 0.37  & 84.00 $\pm$ 0.25 & 60.01 $\pm$ 0.32 & 77.42 $\pm$ 0.23 \\
    $\quad$+ DA & ResNet-12 & 74.56 $\pm$ 0.34  & 87.61 $\pm$ 0.23 & 64.94 $\pm$ 0.33  & 82.10 $\pm$ 0.23 \\
    $\quad$+ MM & ResNet-12 & \textbf{75.67 $\pm$ 0.34} & \textbf{88.37 $\pm$ 0.23} & \textbf{65.02 $\pm$ 0.32} & \textbf{82.42 $\pm$ 0.23} \\
    \midrule
    MCT & ResNet-12 & 75.80 $\pm$ 0.33 & 89.10 $\pm$ 0.42 & 64.84 $\pm$ 0.33  & 81.45 $\pm$ 0.23 \\
    $\quad$+ MM & ResNet-12 & \textbf{76.00 $\pm$ 0.33} & \textbf{89.54 $\pm$ 0.33} & \textbf{66.37 $\pm$ 0.32} &  \textbf{83.11 $\pm$ 0.22} \\
    \bottomrule
  \end{tabular}
  }
  \vskip -0.05in
  \label{tab:benchmark}
\end{table*}

\subsection{Shot Augmentation for Pre-Trained Models}

In the typical meta-learning framework, data augmentations are used during meta-training but not during test time.  On the other hand, in some transfer learning work, data augmentations, such as horizontal flips, random crops, and color jitter, are used during fine-tuning at test time \citep{chen2019closer}. These techniques enable the network to see more data samples during few-shot testing, leading to enhanced performance. 

We propose shot augmentation (see Section~\ref{Anatomy}) to enlarge the number of few-shot samples during testing, and we also propose a variant in which we additionally train using the same augmentations on support data in order to prepare the meta-learner for this test time scenario. Figure~\ref{fig:shotaug} shows the effect of shot augmentation (using only horizontal flips) on performance for MetaOptNet with ResNet-12 backbone trained with Meta-MaxUp. Shot augmentation consistently improves results across datasets, especially on 1-shot classification ($\sim$ 2\%).  To be clear, in this figure, we are not using shot augmentation during the training stage.  Rather, we are using conventional low-shot training, and then deploying our models with shot augmentation at test time.  These post-training performance gains can be achieved by directly applying shot augmentation to pre-trained/existing models during testing.  For additional experiments, see Appendix G.

\subsection{Improving Existing Meta-Learners with Better Data Augmentation}

In this section, we improve the performance of four different popular meta-learning methods including ProtoNet \citep{snell2017prototypical}, R2-D2 \citep{bertinetto2018meta}, MetaOptNet \citep{lee2019meta}, and MCT \citep{kye2020transductive}.  We compare their baseline performance to query CutMix with task-level rotation as well as Meta-MaxUp data augmentation strategies on both the CIFAR-FS and mini-ImageNet datasets.  See Table~\ref{tab:benchmark} for the results of these experiments.  In all cases, we are able to improve the performance of existing methods, sometimes by over 5\%.  Even without Meta-MaxUp, we improve performance over the baseline by a large margin.  The superiority of meta-learners that use these augmentation strategies suggests that data augmentation is critical for these popular algorithms and has largely been overlooked.      

In addition, we compare our method to augmentation by Large Rotations at the task level -- the only competing work to our knowledge -- in Table~\ref{tab:SOTA}.  Note that using Large Rotations to create new classes is referred to as ``Task Augmentation'' in~\citep{liu2020task}; we refer to it here as ``Large Rotations'' to avoid confusion since we study a myriad of augmentations at the task level.  We observe that with the same training algorithm (MetaOptNet with SVM) and the ResNet-12 backbone, our method outperforms the Large Rotations augmentation strategy by a large margin on both the CIFAR-FS and mini-ImageNet datasets. Together with the same ensemble method as used in Large Rotations, marked by ``+ens", we further boost performance consistently above the MCT baseline, the current highest performing meta-learning method on these benchmarks, despite using an older meta-learner previously thought to perform worse than MCT. Moreover, when both training and validation datasets are used for meta-training, we can achieve the state-of-art results for few-shot classification on mini-ImageNet in inductive setting. 

\begin{table*}[h]
  \caption{Few-shot classification accuracy (\%) on CIFAR-FS and mini-ImageNet with ResNet-12 backbone. ``M-SVM" denotes MetaOptNet with the SVM head. ``+ens" denotes testing with ensemble methods as in \cite{liu2020task}. ``LargeRot'' denotes task-level augmentation by Large Rotations as described in \cite{liu2020task}.}
  \vskip 0.15in
  \centering
  \begin{tabular}{@{}lcccc@{}}
    \toprule
    &\multicolumn{2}{c}{ CIFAR-FS}&\multicolumn{2}{c}{mini-ImageNet}\\
    \cmidrule(r){2-3}\cmidrule(r){4-5}
    Method & 1-shot & 5-shot & 1-shot & 5-shot\\
    \midrule
    M-SVM + LargeRot & 72.95 $\pm$ 0.24 & 85.91 $\pm$ 0.18  & 62.12 $\pm$ 0.22 & 78.90 $\pm$ 0.17  \\
     M-SVM + MM (ours) & 75.67 $\pm$ 0.34 & 88.37 $\pm$ 0.23 & 65.02 $\pm$ 0.32 & 82.42 $\pm$ 0.23 \\
    M-SVM + LargeRot + ens & 75.85 $\pm$ 0.24 & 87.73 $\pm$ 0.17 & 64.56 $\pm$ 0.22 & 81.35 $\pm$ 0.16 \\
    M-SVM + MM + ens (ours) & 76.38 $\pm$ 0.33 & 89.16 $\pm$ 0.22  & 66.42 $\pm$ 0.32 & 83.69 $\pm$ 0.21 \\
    M-SVM + LargeRot + ens + val & \textbf{76.75 $\pm$ 0.23} & 88.38 $\pm$ 0.17 & 65.38 $\pm$ 0.23  & 82.13 $\pm$ 0.16 \\
    M-SVM + MM + ens + val (ours) & 76.38 $\pm$ 0.34 & \textbf{89.25 $\pm$ 0.21} & \textbf{67.37 $\pm$ 0.32} & \textbf{84.57 $\pm$ 0.21} \\
    \bottomrule
  \end{tabular}
  \label{tab:SOTA}
\end{table*}

\subsection{Out-of-Distribution Testing on Meta-Dataset}
\label{sec:experiment:metadataset}
In this section, we examine the effectiveness of our methods on cross-domain few-shot learning benchmarks. Few-shot learners may be successful on similar tasks to their training data but fail on tasks that deviate.  Thus, testing on diverse distributions is crucial.  To this end, we leverage Meta-Dataset, a collection of subdatasets used for testing meta-learners across diverse tasks \citep{triantafillou2019meta}.  Among the 10 subdatasets, we train the networks only on ILSVRC-2012~\citep{russakovsky2015imagenet}, the largest dataset in the collection, and we evaluate the cross-domain few-shot classification performance on the other 9 datasets with R2-D2 and MetaOptNet learners and ResNet-12 backbones. Training and evaluation details can be found in Appendix H. 

We observe that on all subdatasets except for Omniglot, our proposed methods can improve test accuracy over the baseline by as much as 7\%. Additionally, we improve performance by a large margin (more than 3\%) on more than half of the subdatasets. On average, Meta-MaxUp improves accuracy by around 3\%. Omniglot suffers under our strategies since this dataset comprises handwritten letters which are not invariant to strong augmentations.  Specially designed augmentations for handwritten letters are necessary to optimize performance on Omniglot.  The success of Meta-MaxUp on cross-domain benchmarks demonstrates that the proposed strategy is effective even on diverse testing distributions which do not resemble the learner's training data.

\begin{table}[h!]
\caption{Few-shot classification accuracy (\%) on Meta-Dataset with both MetaOptNet and R2-D2 learner. ``+ DA" denotes training with CutMix (Q) + Rotation (T), and ``+ MM" denotes training with Meta-MaxUp. Confidence intervals have radius equal to one standard error.}
\vskip 0.15in
\begin{center}
\scalebox{0.88}{
\begin{tabular}{lccc}
\hline
Test Source & R2-D2 & + DA & + MM  \\ \hline
ILSVRC & 69.04 $\pm$ 0.31 & 70.30 $\pm$ 0.31 & \textbf{71.68 $\pm$ 0.30} \\ 
Birds & 75.22 $\pm$ 0.30 & 77.27 $\pm$ 0.28 & \textbf{77.95 $\pm$ 0.30} \\ 
Omniglot & \textbf{97.46 $\pm$ 0.08} & 96.10 $\pm$ 0.11 & 96.71 $\pm$ 0.09\\ 
Aircraft & 54.28 $\pm$ 0.28 & 58.93 $\pm$ 0.30 & \textbf{60.83 $\pm$ 0.28} \\ 
Textures & 63.47 $\pm$ 0.24 & 65.98 $\pm$ 0.24 & \textbf{67.34  $\pm$ 0.26} \\ 
Quick Draw & 76.39 $\pm$ 0.27 & 78.44 $\pm$ 0.27 & \textbf{80.83 $\pm$ 0.25} \\ 
Fungi & 50.41 $\pm$ 0.22 & 52.29 $\pm$ 0.20 & \textbf{54.12 $\pm$ 0.22} \\ 
VGG Flower & 86.26 $\pm$ 0.21 & 87.79 $\pm$ 0.19 & \textbf{90.29 $\pm$ 0.17} \\ 
Traffic Signs & 83.98 $\pm$ 0.34 & \textbf{84.23 $\pm$ 0.36} & 83.59 $\pm$ 0.36 \\ 
MSCOCO & 70.29 $\pm$ 0.30 & 71.59 $\pm$ 0.31 & \textbf{72.83 $\pm$ 0.29} \\ \hline
\vspace{-0.2cm}\\\hline
Test Source & MetaOptNet  & + DA & + MM  \\ \hline
ILSVRC & 68.92 $\pm$ 0.30 & 71.17 $\pm$ 0.30 & \textbf{72.19 $\pm$ 0.30} \\ 
Birds & 75.58 $\pm$ 0.39 & \textbf{77.49 $\pm$ 0.29} & 77.47 $\pm$ 0.2 \\ 
Omniglot & \textbf{97.43  $\pm$ 0.10} & 95.97 $\pm$ 0.10 & 96.59 $\pm$ 0.09\\ 
Aircraft & 53.40 $\pm$ 0.37 & 60.43 $\pm$ 0.29 & \textbf{60.57 $\pm$ 0.29} \\ 
Textures & 63.29 $\pm$ 0.33 & 65.70 $\pm$ 0.24 & \textbf{69.42 $\pm$ 0.25} \\ 
Quick Draw & 78.00  $\pm$ 0.33 & 79.56 $\pm$ 0.25 & \textbf{80.67 $\pm$ 0.25} \\ 
Fungi & 50.56 $\pm$ 0.21 & 53.80 $\pm$ 0.22 & \textbf{53.82 $\pm$ 0.22}  \\ 
VGG Flower & 88.16 $\pm$ 0.25 & 89.92 $\pm$ 0.18 & \textbf{91.13 $\pm$ 0.15} \\ 
Traffic Signs & 85.12 $\pm$ 0.33 & \textbf{85.25 $\pm$ 0.33} & 83.38 $\pm$ 0.37 \\ 
MSCOCO & 69.52 $\pm$ 0.32 & 71.90 $\pm$ 0.31 & \textbf{73.49 $\pm$ 0.30} \\ \hline

\end{tabular}
}
\end{center}
\label{tab:metadataset-r2d2}
\vskip -0.1in
\end{table}

\section{Discussion}
\label{Discussion}

In this work, we break down data augmentation in the context of meta-learning.  In doing so, we uncover possibilities that do not exist in the classical image classification setting. We identify four modes of augmentation: query, support, task, and shot. These modes behave differently and are of varying importance. Specifically, we find that augmenting query data is particularly important.  After adapting various data augmentations to meta-learning, we propose Meta-MaxUp for combining various meta-specific data augmentations.  We demonstrate that Meta-MaxUp significantly improves the performance of popular meta-learning algorithms. As shown by the recent popularity of frameworks like AutoAugment~\cite{cubuk2018autoaugment} and MaxUp~\cite{gong2020maxup}, data augmentation for standard classification is still an active area of research.  We hope that this work opens up possibilities for further work on meta-specific data augmentation and that emerging methods for data augmentation will boost the performance of meta-learning on progressively larger models with more complex backbones. 

\section*{Acknowledgement}
This work was supported by the AFOSR MURI program, the Office of Naval Research, the DARPA YFA program, and the National Science Foundation Directorate of Mathematical Sciences.  Additional support was provided by Capital One Bank and JP Morgan Chase.


\bibliography{main}
\bibliographystyle{icml2021}

\clearpage
\appendix
\onecolumn

\section{Impact of Dataset Diversity on Various Stages of Meta-learning}
\label{appendix:random_gen}

Following the settings in Section~\ref{data_size}, we further investigate how dataset diversity matters in the various stages of meta-learning on Mini-ImageNet, CNN-4 backbones, and with the ProtoNet~\cite{snell2017prototypical} head. 
The results are shown in Table~\ref{tab:data_size_more}, and all these results support our findings that meta learning algorithms are more sensitive to the amount of query data and number of tasks and less sensitive to the amount of support data.

\begin{table}[h!]
\caption{Few-shot classification accuracy (\%) using different meta-learning algorithms and backbones for various data size manipulations on Mini-ImageNet. ``Support'', ``Query'' and ``Task'' columns denote the number of samples per class for support and query data and the number of total tasks available for sampling. 
Confidence intervals have radius equal to one standard error.}
\vskip 0.15in
\begin{center}
\begin{tabular}{lccccccc}
\hline
Method & Backbone & Support & Query & Task & 1-shot & 5-shot \\ 
\midrule
R2-D2& CNN-4 & 600 & 600 & full & 55.94 $\pm$ 0.32 & 72.32 $\pm$ 0.25 \\ 
R2-D2& CNN-4 & 5 & 600 & full & 55.05 $\pm$ 0.31 & 71.66 $\pm$ 0.26 \\ 
R2-D2& CNN-4 & 5 (random) & 600 & full & 42.09 $\pm$ 0.29 & 60.12 $\pm$ 0.27\\ 
R2-D2& CNN-4 & 600 & 5 & full & 49.87 $\pm$ 0.30 & 66.00 $\pm$ 0.27 \\ 
R2-D2& CNN-4 & 600 & 600 & 13 & 53.34 $\pm$ 0.31 & 69.43 $\pm$ 0.25 \\ 
\midrule
R2-D2& ResNet-12 & 600 & 600 & full & 60.50 $\pm$ 0.33 & 76.60 $\pm$ 0.24 \\ 
R2-D2& ResNet-12 & 5 & 600 & full & 58.79 $\pm$ 0.32 & 76.45 $\pm$ 0.25 \\ 
R2-D2& ResNet-12 & 5 (random) & 600 & full & 43.80 $\pm$ 0.30 & 62.26 $\pm$ 0.28\\ 
R2-D2& ResNet-12 & 600 & 5 & full & 48.02 $\pm$ 0.31 & 65.45 $\pm$ 0.26 \\ 
R2-D2& ResNet-12 & 600 & 600 & 13 & 57.65 $\pm$ 0.33 & 73.18 $\pm$ 0.28 \\ 
\midrule

ProtoNet & ResNet-12 & 600 & 600 & full & 57.46 $\pm$ 0.38 & 75.61 $\pm$ 0.29 \\ 
ProtoNet & ResNet-12 & 5 & 600 & full & 57.03 $\pm$ 0.33 & 75.46 $\pm$ 0.26 \\ 
ProtoNet & ResNet-12 & 5 (random) & 600 & full & 43.36 $\pm$ 0.32 & 57.20 $\pm$ 0.28\\ 
ProtoNet & ResNet-12 & 600 & 5 & full & 48.40 $\pm$ 0.34 & 64.79 $\pm$ 0.29 \\ 
ProtoNet & ResNet-12 & 600 & 600 & 13 & 51.88 $\pm$ 0.35 & 66.41 $\pm$ 0.29 \\ \hline
\end{tabular}
\end{center}
\label{tab:data_size_more}
\end{table}

\section{Details About Data Augmentation Techniques}
\label{appendix:techniques}

In this section, we provide more details about the different data augmentation techniques we use in this work. We employ the following pool of data augmentations:

\paragraph{CutMix:} \citet{yun2019cutmix} introduce the CutMix augmentation strategy where patches are cut and pasted among training images, and the ground truth labels are also mixed proportionally to the area of the patches.

\paragraph{MixUp:} \citet{zhang2017mixup} propose mixup, a simple learning principle to alleviate memorization and sensitivity to adversarial examples. Mixup trains a neural network on convex combinations of pairs of examples and their labels. By doing so, mixup regularizes the neural network to favor simple linear behavior in between training examples.

\paragraph{Self-Mix:} \citet{seo2020self} introduce the self-mix augmentation strategy in which a patch of an image is substituted into other values in the same image to improve the generalization ability of few-shot image classification models.

In addition, we use some standard and simple data augmentation techniques:

\paragraph{Rotation:} augments the data by rotating the images. 

\paragraph{Horizontal Flip:} augments the data by horizontally flipping images.

\paragraph{Random Erase:} augments the data by randomly erasing patches from the image.

Finally, we also experimented with the following data augmentation techniques:

\paragraph{Combining Labels:} augments the data by combining two different labels into a single class. For instance, we may combine the ``dog'' and ``cat'' labels to create a new ``dog or cat'' class.

\paragraph{Feature Mixup:} similar to the ``Mixup'' augmentation technique we describe above, however we perform the mixup strategy on the feature representation for the image.

\paragraph{Drop Channel:} augments the data by dropping color channels in the image.

\paragraph{Solarize:} inverts all pixels above a threshold value of magnitude.

\section{Training Details}
\label{appendix:traindetail}
For MetaOptNet, we use the same training procedure as~\citep{lee2019meta} including SGD with Nesterov momentum of 0.9 and weight decay coefficient 0.0005. The model was meta-trained for 60 epochs, with an initial learning rate 0.1, then changed to 0.006, 0.0012, and 0.00024 at epochs 20, 40 and 50, respectively. In each epoch, we train on 8000 episodes and use mini-batches of size 8.  Following~\citep{lee2019meta}, we use a larger shot number (15) to train mini-ImageNet for both 1-shot and 5-shot classification. For MCT, we use the same optimizer but with batch size 1 and maximum iterations 50000. Following~\citep{kye2020transductive}, we enlarge the training  classification ways to 15 for a 5-way testing. We use instance-wise metric for all inductive learning. 

Table~\ref{tab:training} compares the training time of meta-learning methods with baseline data augmentations, with our proposed data augmentations (DA) and Meta-MaxUp strategy (MM) on the CIFAR-FS dataset. We employ data parallelism across 4 Nvidia RTX 2080 Ti GPUs for all experiments. The training time of meta algorithms with our proposed data augmentation is almost the same as with baseline methods. Although Meta-MaxUp requires $m$ times as many forward passes, here $m=4$, it does not require any extra backward passes. Thus, Meta-MaxUp typically runs roughly 2-3 times longer than baseline methods.

\begin{table}[h!]
\caption{Runtime (training time in hours for 60 epochs) comparison of data augmentation strategies on CIFAR-FS}
\vskip 0.15in
\begin{center}
\begin{tabular}{lcccc}
\hline
Method & Backbone & Runtime & Backbone & Runtime \\ \hline
R2D2 & CNN-4 & 2.5h &  ResNet-12 & 3.2h \\ \hline
\quad + DA & CNN-4 & 2.6h  & ResNet-12 & 3.7h \\ \hline
\quad + MM & CNN-4 & 4.1h  & ResNet-12 & 8.2h \\ \hline
MetaOptNet & CNN-4 & 8.6h & ResNet-12 & 8.9h \\ \hline
\quad + DA & CNN-4 & 8.8h & ResNet-12 & 9.2h\\ \hline
\quad + MM & CNN-4 & 14.5h & ResNet-12 & 18.5h \\ \hline
\end{tabular}

\end{center}
\label{tab:training}
\end{table}

\section{Results for All Data Augmentation Techniques}
Table~\ref{tab:fulltable} shows the few-shot classification accuracy on CIFAR-FS of an R2-D2 meta-learner with all single data augmentation techniques used in our paper. We highlight the best result in each mode. Data augmentation on query images significantly improves the baseline performance as well as data augmentations on other modes.

\label{appendix:full_table}
\begin{table}[!h]
  \caption{Few-shot classification accuracy (\%) on the CIFAR-FS dataset for all  data augmentations with an R2-D2 learner. Confidence intervals have radius equal to one standard error. ``CNN-4'' denotes a 4-layer convolutional network with 96, 192, 384, and 512 filters in each layer~\citep{bertinetto2018meta}.  Best performance in each category is bolded.}
  \vskip 0.1in
  \centering
  \begin{tabular}{@{}lccccc@{}}
    \toprule
    &&\multicolumn{2}{c}{ CNN-4}&\multicolumn{2}{c}{ResNet-12}\\
    \cmidrule(r){3-4}\cmidrule(r){5-6}
    Mode &Level & 1-shot & 5-shot & 1-shot & 5-shot\\
    \midrule
    Baseline & - & 67.56 $\pm$ 0.35 & 82.39 $\pm$ 0.26  &71.95 $\pm$ 0.37 & 84.56 $\pm$ 0.25  \\
    \midrule
    Random Erase & Support & 67.71 $\pm$ 0.36 & 82.25 $\pm$ 0.26 & 72.30 $\pm$ 0.37 & 84.50 $\pm$ 0.25 \\
    Self-Mix & Support & \textbf{69.61 $\pm$ 0.35} & \textbf{83.43 $\pm$ 0.25} & 71.96 $\pm$ 0.36 & \textbf{84.84 $\pm$ 0.25} \\
    CutMix& Support & 69.05 $\pm$ 0.36 & 83.12 $\pm$ 0.26  & \textbf{72.79 $\pm$ 0.37} & 84.70 $\pm$ 0.25 \\
    MixUp & Support & 68.64 $\pm$ 0.37 & 82.72 $\pm$ 0.27 & 71.86 $\pm$ 0.37 & 84.11 $\pm$ 0.25 \\
    Feature Mixup & Support & 67.88 $\pm$ 0.35 & 82.40 $\pm$ 0.25 & 71.21 $\pm$ 0.37  &83.38 $\pm$ 0.25 \\
    Rotation & Support & 68.65 $\pm$  0.35 & 82.86 $\pm$ 0.25 & 71.13 $\pm$ 0.37 & 83.84 $\pm$ 0.25 \\
    Combining labels & Support & 68.27 $\pm$ 0.36 & 82.53 $\pm$ 0.26 & 71.00 $\pm$ 0.38 & 83.12 $\pm$ 0.25 \\
    Drop Channel & Support & 68.21 $\pm$ 0.35 & 82.76 $\pm$ 0.25 & 69.65 $\pm$ 0.73 & 83.15 $\pm$ 0.25 \\
    Solarize & Support & 68.65 $\pm$ 0.35 & 82.68 $\pm$ 0.26 & 70.88 $\pm$ 0.37 & 83.45 $\pm$ 0.25 \\
    \midrule
    Random Erase & Query & 69.73 $\pm$ 0.34 & 84.04 $\pm$ 0.25 & 73.05 $\pm$ 0.36 & 85.67 $\pm$ 0.25 \\
    Self-Mix & Query & 69.55 $\pm$ 0.35 & 84.20 $\pm$ 0.25 &73.59 $\pm$ 0.35 & 86.14 $\pm$ 0.25 \\
    CutMix & Query & \textbf{70.54 $\pm$ 0.33} & \textbf{84.69 $\pm$ 0.24} & \textbf{75.97 $\pm$ 0.34} & \textbf{87.28 $\pm$ 0.23} \\
    MixUp & Query & 67.70 $\pm$ 0.34 & 83.13 $\pm$ 0.25 & 72.93 $\pm$ 0.35 & 86.13 $\pm$ 0.24 \\
    Feature Mixup & Query & 70.16 $\pm$ 0.35 & 83.80 $\pm$ 0.28 & 73.38 $\pm$ 0.35 & 85.87 $\pm$ 0.23 \\
    Rotation & Query & 68.17 $\pm$ 0.35 & 83.01 $\pm$ 0.25 & 72.02 $\pm$ 0.36 & 84.42 $\pm$ 0.25 \\
    Combining labels & Query & 66.01 $\pm$ 0.34 & 81.99 $\pm$ 0.26 & 69.77 $\pm$ 0.37 & 82.99 $\pm$ 0.26 \\
    Drop Channel & Query & 68.34 $\pm$ 0.35 & 83.25 $\pm$ 0.25 & 69.60 $\pm$ 0.37 & 83.01 $\pm$ 0.26 \\
    Solarize & Query &67.51 $\pm$ 0.35 & 82.65 $\pm$ 0.25 & 72.45 $\pm$ 0.36 & 84.97 $\pm$ 0.24 \\
    \midrule
    MixUp & Task & 67.21 $\pm$ 0.35 & 82.72 $\pm$ 0.26 & 72.05 $\pm$ 0.37 & 85.27 $\pm$ 0.25 \\
    Large Rotation & Task & \textbf{68.96 $\pm$ 0.35} & \textbf{83.65 $\pm$ 0.25} & \textbf{73.79 $\pm$ 0.36 }& \textbf{85.81 $\pm$ 0.24} \\
    CutMix & Task & 68.78 $\pm$ 0.36 & 82.99 $\pm$ 0.50 & 72.72 $\pm$ 0.37 & 84.62 $\pm$ 0.25  \\
    Combining labels & Task & 68.08 $\pm$ 0.35 &82.33 $\pm$ 0.26 &69.64 $\pm$ 0.37 & 83.79 $\pm$ 0.26 \\
    Random Erase & Task & 68.39 $\pm$ 0.36 & 83.26 $\pm$ 0.25 & 71.09 $\pm$ 0.37 & 84.49 $\pm$ 0.25\\
    Drop Channel & Task & 67.54 $\pm$ 0.36 & 81.97 $\pm$ 0.25 & 70.24 $\pm$ 0.37 & 83.52 $\pm$ 0.26 \\
    \midrule
    Horizontal Flip & Shot & \textbf{68.13 $\pm$ 0.35} & 82.95 $\pm$ 0.25 & \textbf{73.25 $\pm$ 0.36} & \textbf{85.06 $\pm$ 0.25} \\
    Random Crop & Shot & 67.33 $\pm$ 0.36 & \textbf{83.04 $\pm$ 0.25} & 70.56 $\pm$ 0.37 & 83.87 $\pm$ 0.25 \\
    Random Rotation & Shot & 67.57 $\pm$ 0.35 & 83.00 $\pm$ 0.25 & 70.32 $\pm$ 0.37 & 83.75 $\pm$ 0.25 \\
    \bottomrule
  \end{tabular}
  \label{tab:fulltable}
\end{table}

\section{Results for Combination of Data Augmentations}
Table~\ref{tab:combination_full} shows the few-shot classification accuracy for combinations of data augmentations building on the top of query CutMix, with both CNN-4 and ResNet-12 backbones.
\label{appendix:combine}
\begin{table}[!h]
    \caption{Few-shot classification accuracy (\%) on the CIFAR-FS dataset with combinations of augmentations and query CutMix. ``S",``Q",``T" denote ``Support", ``Query", and ``Task" modes, respectively.  While adding augmentations can help, it can also hurt, so additional augmentations must be chosen carefully.}
  \vskip 0.1in
  \centering
  \begin{tabular}{@{}lcccc@{}}
    \toprule
    &\multicolumn{2}{c}{ CNN-4}&\multicolumn{2}{c}{ResNet-12}\\
    \cmidrule(r){2-3}\cmidrule(r){4-5}
    Mode & 1-shot & 5-shot & 1-shot & 5-shot\\
    \midrule
    CutMix & 70.54 $\pm$ 0.33 & 84.69 $\pm$ 0.24  &75.97 $\pm$ 0.34 & 87.28 $\pm$ 0.23  \\
    + CutMix (S) & 69.50 $\pm$ 0.35 & 82.64 $\pm$ 0.26 & 75.00 $\pm$ 0.37 & 85.37 $\pm$ 0.25 \\
    + Random Erase (S) & 70.12 $\pm$ 0.35 & 84.48 $\pm$ 0.25 & 75.84 $\pm$ 0.34 & 87.19 $\pm$ 0.24 \\
    + Random Erase (Q) & 69.68 $\pm$ 0.34 & 84.36 $\pm$ 0.24 & 75.08 $\pm$ 0.35 & 87.14 $\pm$ 0.23  \\
    + Self-Mix (S) & 70.65 $\pm$ 0.34 & 84.68 $\pm$ 0.25 & \textbf{76.27 $\pm$ 0.34} & 87.52 $\pm$ 0.24 \\
    + Self-Mix (Q) & 69.94 $\pm$ 0.34 & 84.38 $\pm$ 0.24 & 76.04 $\pm$ 0.34 & 87.45 $\pm$ 0.24 \\
    + MixUp (T) & 70.33 $\pm$ 0.35 & 84.57 $\pm$ 0.25 & 75.97 $\pm$ 0.34  & 86.66 $\pm$ 0.24 \\
    + Rotation (T) & 70.35 $\pm$ 0.34 & 84.73 $\pm$ 0.24 & 75.74 $\pm$ 0.34 & \textbf{87.68 $\pm$ 0.24} \\
    + Horizontal Flip (Shot) & \textbf{70.90 $\pm$ 0.33} & \textbf{84.87 $\pm$ 0.24}  & 76.23 $\pm$ 0.34 & 87.36 $\pm$ 0.24 \\
    \bottomrule
  \end{tabular}
  \label{tab:combination_full}
\end{table}

\clearpage
\section{Augmentation Pool for Meta-MaxUp}
\label{appendix:maxup}

For all the benchmark results of Meta-MaxUp training, we use a medium-size data augmentation pool with $m = 4$, including CutMix (Q), Random Erase (Q), Self-Mix (S), Rotation (T), CutMix (Q) + Rotation (T), and Random Erase (Q) + Rotation (T). For the large-size pool, we add more techniques and combinations of the mentioned techniques into the pool, including Random Erase (Q) + Random Erase (S), CutMix (Q) + Random Erase (S), CutMix (Q) + Random Erase (Q), and CutMix (Q) + Self-Mix (S). Table~\ref{tab:maxup_fulltable} shows the few-shot classification accuracy on CIFAR-FS using R2D2 meta-learner and both CNN-4, ResNet-12 backbones, with various augmentations pool and hyper-parameter $m$.

\begin{table}[!h]
    \caption{Few-shot classification accuracy (\%) on the CIFAR-FS dataset for Meta-MaxUp over different sizes of augmentation pools and numbers of samples.  As $m$ and the pool size increase, so does performance.  Meta-MaxUp is able to pick effective augmentations from a large pool.}
    \vskip 0.15in
  \centering
  \begin{tabular}{@{}lccccc@{}}
    \toprule
    & &\multicolumn{2}{c}{ CNN-4}&\multicolumn{2}{c}{ResNet-12}\\
    \cmidrule(r){3-4}\cmidrule(r){5-6}
    Pool & m & 1-shot & 5-shot & 1-shot & 5-shot\\
    \midrule
    Baseline & - & 67.56 $\pm$ 0.36 & 82.39 $\pm$ 0.26  &71.95 $\pm$ 0.37 & 84.56 $\pm$ 0.25  \\
    \midrule
    CutMix & 1 & 70.54 $\pm$ 0.34 & 84.69 $\pm$ 0.24 & 75.97 $\pm$ 0.34 & 87.28 $\pm$ 0.23 \\
    Single & 1 & 70.76 $\pm$ 0.35 & 84.70 $\pm$ 0.25 & 75.71 $\pm$ 0.35 & 87.44 $\pm$ 0.43  \\
    Medium & 1 & 70.50 $\pm$ 0.34 & 84.59 $\pm$ 0.24 & 75.60 $\pm$ 0.34 & 87.35 $\pm$ 0.23 \\
    Large & 1 & 70.84 $\pm$ 0.34 & 85.04 $\pm$ 0.24 & 75.44 $\pm$ 0.34 & 87.47 $\pm$ 0.23  \\
    \midrule
    CutMix & 2 &  70.56 $\pm$ 0.34 & 84.78 $\pm$ 0.24 & 74.93 $\pm$ 0.36 & 87.14 $\pm$ 0.24 \\
    Single & 2 & 70.86 $\pm$ 0.34 & 85.06 $\pm$ 0.25 & 75.81 $\pm$ 0.34 & 87.33 $\pm$ 0.23 \\
    Medium & 2 & 70.75 $\pm$ 0.34 & 85.02 $\pm$ 0.24 & 76.49 $\pm$ 0.33 & 88.20 $\pm$  0.22 \\
    Large & 2 & 70.63 $\pm$ 0.34 & 85.07 $\pm$ 0.24 & 76.59 $\pm$ 0.34 &  88.11 $\pm$ 0.23\\
    \midrule
    CutMix & 4 & 70.48 $\pm$ 0.34 & 84.76 $\pm$ 0.24 & 75.08 $\pm$ 0.23 & 87.60 $\pm$ 0.24 \\
    Single & 4 & \textbf{71.10 $\pm$ 0.34} & \textbf{85.50 $\pm$  0.24} & 76.82 $\pm$ 0.24 &  88.14 $\pm$ 0.23\\
    Medium & 4 & 70.58 $\pm$ 0.34 & 85.32 $\pm$ 0.24 & 76.30 $\pm$ 0.24 &  88.29 $\pm$  0.22\\
    Large & 4 & 70.71 $\pm$ 0.34  & 85.04 $\pm$ 0.23 & \textbf{76.99 $\pm$ 0.24} & \textbf{88.35 $\pm$  0.22} \\
    \bottomrule
  \end{tabular}
  \label{tab:maxup_fulltable}
\end{table}

\section{Bar Plots for Shot Augmentation}
\label{appendix:moreplots}
\begin{figure}[h]
    \centering
    \subfloat[]{{\includegraphics[scale=0.2]{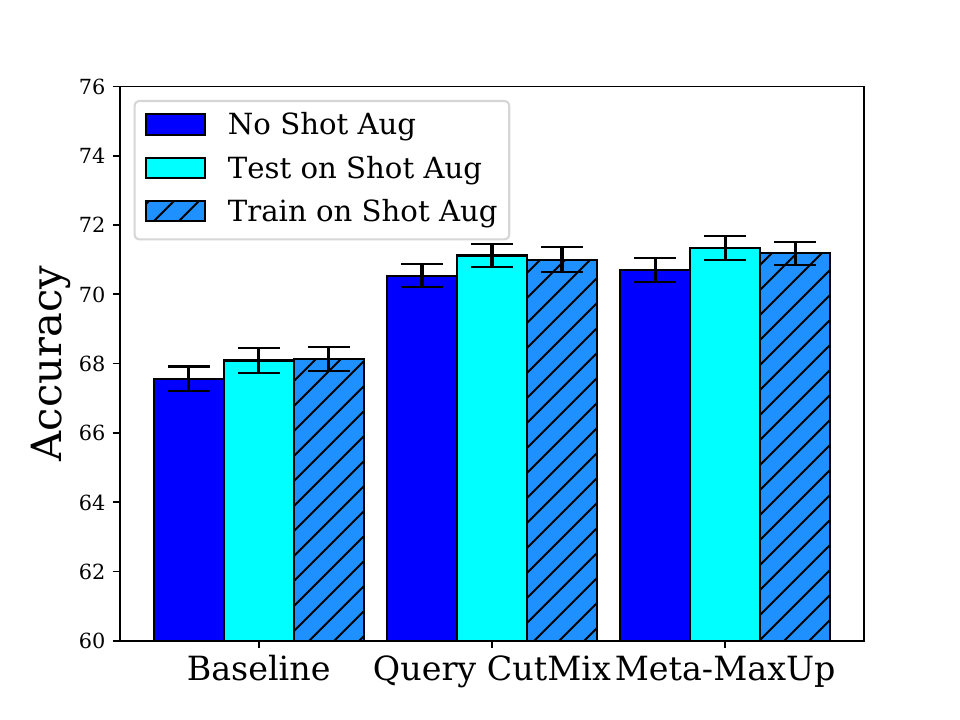}}}
    \subfloat[]{{ \includegraphics[scale=0.2]{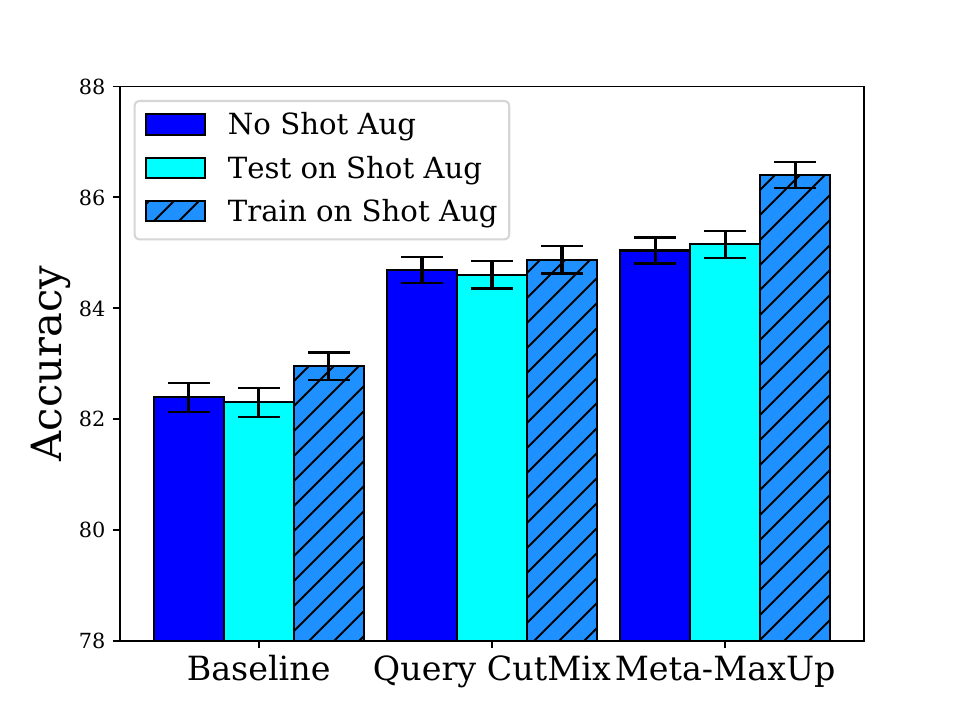} }}
    \subfloat[]{{ \includegraphics[scale=0.2]{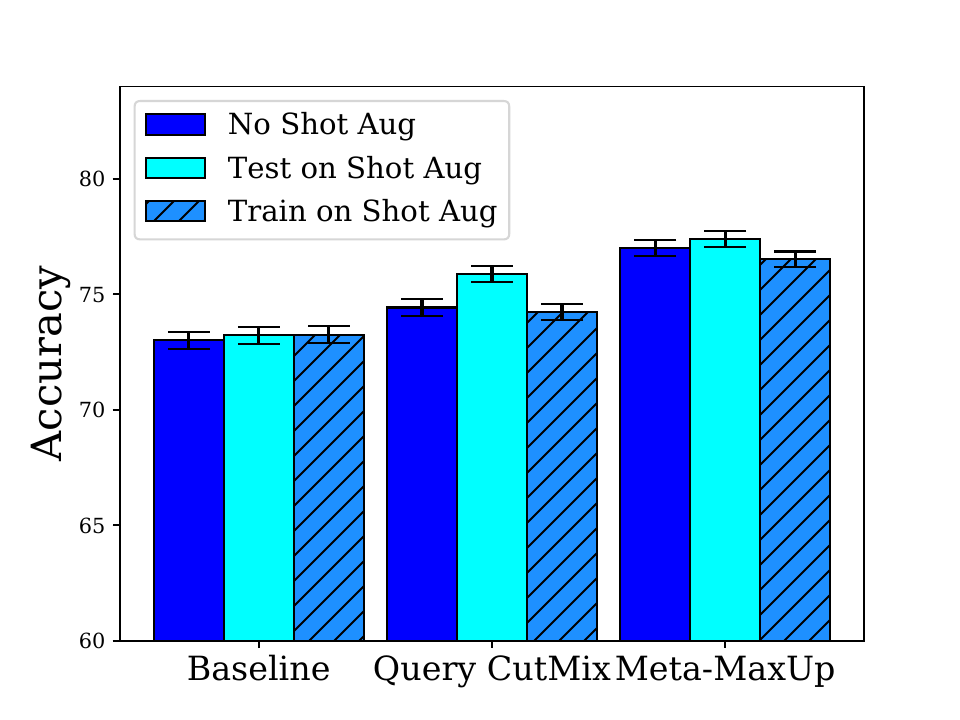} }}
    \subfloat[]{{\includegraphics[scale=0.2]{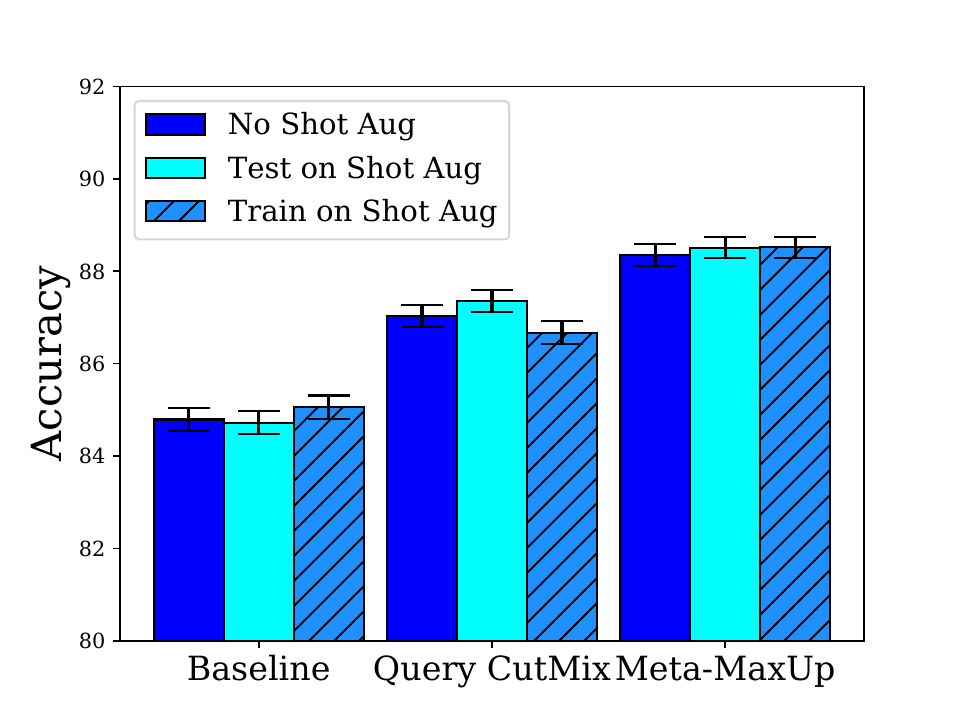} }}
\caption{Performance with shot augmentation using different backbones and training strategies on CIFAR-FS. (a) 1-shot classification with CNN-4 (b) 5-shot classification with CNN-4 (c) 1-shot classification with ResNet-12 (d) 5-shot classification with ResNet-12}\label{fig:moreshotaug}
\end{figure}

Figure~\ref{fig:moreshotaug} shows the effect of the shot augmentation in few-shot evaluation. In general, shot augmentation enhances the performance of meta-learners.

\section{Meta-Dataset Training and Evaluation}
\label{appendix:metadataset}

Details concerning each dataset in the Meta-Dataset benchmark can be found in ~\citet{triantafillou2019meta}. We use the same training procedure as mentioned above in Appendix~\ref{appendix:traindetail} for both meta-learners. In each epoch, we train on 8000 episodes with shot of 5 and images of spacial dimensions $84 \times 84 \times 3$, and we use mini-batches of 8 tasks each. When training with Meta-MaxUp, we use the same augmentation pool as in Appendix~\ref{appendix:maxup} and set $m=4$. During evaluation, we test 5-shot performance on 1000 tasks consisting of 15 query samples each. Due to the small number of sample size for several classes in the Fungi dataset, we use 1-shot classification with 5 query samples instead.

\end{document}